# Modeling emotion for human-like behavior in future intelligent robots

**Authors:** Marwen Belkaid[1,2] and Luiz Pessoa[3]

**Affiliations**
[1] Istituto Italiano di Tecnologia (IIT), Genova, Italy.
[2] Sorbonne Université, CNRS, Institut des Systèmes Intelligents et de Robotique (ISIR), 75005 Paris, France.
[3] Maryland Neuroimaging Center and Departments of Psychology and Electrical and Computer Engineering, University of Maryland, College Park, USA

[*] **Corresponding author:**
Address:
Istituto Italiano di Tecnologia
Center for Human Technologies
Via Enrico Melen, 83
16152 Genoa, Italy
Phone:
+39 010 817 2246

**Email addresses**
marwen.belkaid@iit.it  (Marwen Belkaid), pessoa@umd.edu (Luiz Pessoa)

**Abstract**
Over the past decades, research in cognitive and affective neuroscience has emphasized that emotion is crucial for human intelligence and in fact inseparable from cognition. Concurrently, there has been growing interest in simulating and modeling emotion-related processes in robots and artificial agents. In this opinion paper, our goal is to provide a snapshot of the present landscape in emotion modeling and to show how neuroscience can help advance the current state of the art. We start with an overview of the existing literature on emotion modeling in three areas of research: affective computing, social robotics, and neurorobotics. Briefly summarizing the current state of knowledge on natural emotion, we then highlight how existing proposals in artificial emotion do not make sufficient contact with neuroscientific evidence. We conclude by providing a set of principles to help guide future research in artificial emotion and intelligent machines more generally. Overall, we argue that a stronger integration of emotion-related processes in robot models is critical for the design of human-like behavior in future intelligent machines. Such integration not only will contribute to the development of autonomous social machines capable of tackling real-world problems but would contribute to advancing understanding of human emotion.

In recent decades there has been increased interest in modeling *emotion* in robots and artificial agents. Proposed attempts at tackling this challenge stem from diverse fields, including social robotics, neurorobotics, and affective computing, which differ in terms of their theoretical foundations, engineering approaches, and research goals. As such, existing approaches are largely disconnected from one another. However, progress in the field requires identifying common threads so that the strengths and weaknesses of different proposals and frameworks can be evaluated. A key goal of the present opinion paper is to outline such a synthesis so as to identify fruitful research directions in the development of emotion in intelligent machines.

But why should machines be concerned with emotion in the first place? In a nutshell, because flexible, intelligent behaviors critically rely on emotion. Although the definition of "emotion" is broad and subject to debate, here we focus on *affective processes* that help determine how we perceive the world, how we learn and remember past experiences, how we make decisions, how we adapt to new situations, and how we communicate with each other. The question of whether machines can experience "feelings" or show "empathy" is, however, beyond the scope of the paper.

Considerable effort has been devoted to developing systems that detect emotions in human users and display simulated emotional expressions in response. However, the brain and behavior literature shows that emotion is involved in nearly all mental processes underlying intelligent behavior, including perception, attention, decision-making, and action. Accordingly, we focus on how to go beyond the recognition/production of emotional expressions so as to allow robots to tackle real-world problems, such as navigating an unknown environment or handling an uncontrolled social interaction. For robots to deal intelligently with such problems, much like in humans, affective processing should be integrated across the computational processes that determine their behavior. Our aim is to bridge the gap between computational models of intelligent behaviors and the neuroscience of emotion, while providing a set of recommendations for future research.

## 1. Virtual and robotic models of emotion

Rather than presenting a comprehensive review of computational and robotic models of emotion, our aim is to provide a snapshot of the present landscape. Thus, based on research in the past two decades in the areas of affective computing, neurorobotics, and social robotics, we selected representative examples[1] according to specific criteria so as to identify common research threads and potential weaknesses.

## 1.1. Snapshot of the existing literature

The present research landscape (Figure 1) can be organized in terms of five criteria: embodiment, behavior, architecture design, theoretical approach, and research goal.

**Embodiment (virtual vs. physical):** Models of emotion can be either virtually or physically embodied. 3D animated conversational agents (Pelachaud, 2009; Courgeon et al., 2011; Sagar et al., 2016), sometimes integrated in virtual reality (Martin et al., 2011; Ochs et al., 2016), can exhibit a rich behavioral repertoire (e.g. gestures, postures, and facial expressions) to convey socio-emotional cues to human users. Expressive (Breazeal, 2003; Karaouzene et al., 2013; Correira et al., 2016; Masuyama et al., 2018) and non-expressive robots (Avila-Garcia & Cañamero 2004; Krichmar, 2013; Belkaid et al., 2018) are physically embodied machines that are situated in the real world and thus have the ability to interact with it and act upon it.

---

1  For further examples of computational models and architectures of emotion, we invite the reader to refer to Marsella and colleagues (2010), Kowalczuk and& Czubenko (2016), and Moerland and colleagues (2018).

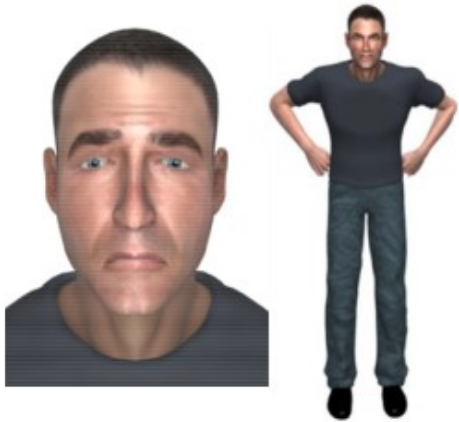
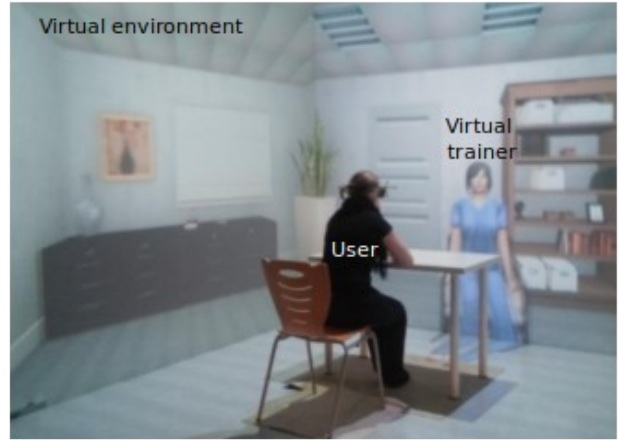
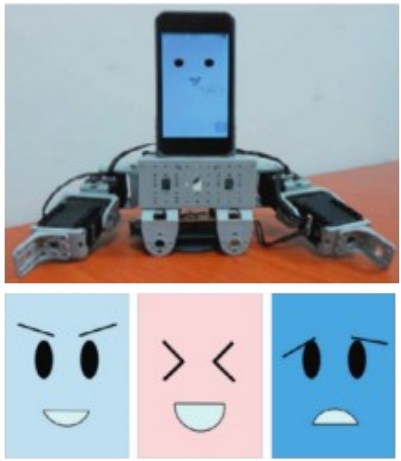
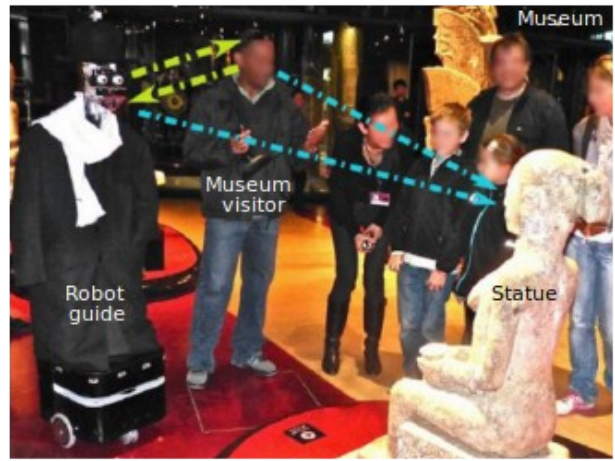
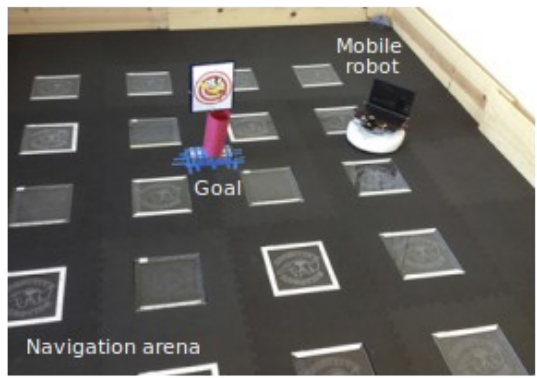
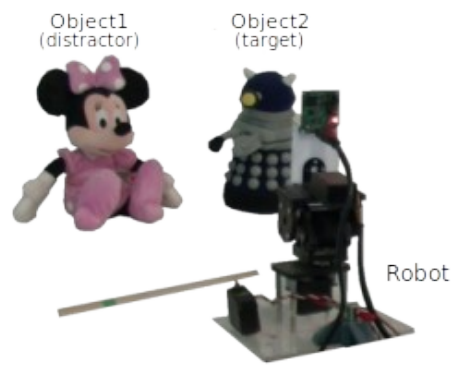

*Figure 1: Emotion modeling approaches. In affective computing, virtual agents convey socio-emotional signals through facial expression and postures (top, left; Courgeon et al., 2011), such as in professional training applications (top, right; Ochs et al., 2016). In social robotics, expressive robots, which may have robotic appearances (middle, left; Masuyama et al., 2018) or anthropomorphic*

*characteristics mimicking human features such as a face with eyes and mouth (middle, right; Karaouzene et al., 2013) are used for situated interactions with humans in applications such as museum guides (middle, right; Karaouzene et al., 2013). In neurorobotics, affective processing helps robots navigate their environments (bottom, left; Krichmar, 2013) and visually explore objects in the visual scene (bottom, right; Belkaid et al., 2017). Pictures are from Courgeon et al., 2011 (top, left), Ochs et al., 2016 (top, right), Masuyama et al., 2018 (middle, left), Karaouzene et al., 2013 (middle, right), Krichmar, 2013 (bottom, left). Pictures from Ochs et al., 2016 (top, right), Karaouzene et al., 2013 (middle, right), Krichmar, 2013 (bottom, left) and Belkaid et al., 2017 (bottom, right) have been annoted by the authors of the present paper.*

**Behavior (social vs. non-social):** Whereas the majority of artificial emotion models focus on social behaviors for the purpose of facilitating human-machine interactions (virtual trainers: Gratch & Marsella, 2004; companion robots: Correira et al., 2016; chatbots: Yacoubi & Sabouret, 2018; erobots: Dubé & Anctil, 2021), another family of models addresses behaviors related to autonomy (foraging: Avila-Garcia & Cañamero, 2004, Krichmar, 2013; visual search: Belkaid et al., 2017). However, few models address both social and non-social behaviors simultaneously (for an exception related to action selection, see Khan & Cañamero, 2018).

**Architecture design (modular vs. integrative):** When models follow a modular design, emotion is simply *added* to the overall system architecture as a separate module which communicates with other components; for example, to activate certain expressive behaviors (e.g. facial expression, posture) in an action module (Breazeal, 2003; Courgeon et al., 2011; Correira et al., 2016). In contrast, integrative approaches highlight the interdependence between emotion and cognition by distributing emotional processing across the system (Krichmar, 2013; Belkaid et al., 2018).

**Theoretical approach (top-down vs. bottom-up):** Taking direct inspiration from emotion theories developed by psychologists[2], some models explicitly instantiate and test theoretical principles in what can be called a top-down fashion (Gratch & Marsella, 2004; Yacoubi & Sabouret, 2018). Bottom-up approaches, however, evaluate behaviors emerging from low-level mechanisms, thus providing concrete implementations of emotion-related processes that are otherwise outlined descriptively (Avila-Garcia & Cañamero, 2004; Boucenna et al., 2014; Sagar et al., 2016).

**Research goal (application-oriented vs. modeling-oriented):** A significant portion of the literature on emotion modeling has aimed to enrich human-machine interactions across multiple applications, such as elderly care (Correira et al., 2016) or training in the context of high-stakes decision-making (Gratch & Marsella, 2004). A complementary goal is to model mechanisms of natural emotion to advance the understanding of the neural systems involved (Balkenius et al., 2009; Krichmar, 2013; Belkaid et al., 2018; Khan & Cañamero, 2018; Lowe et al., 2019). In this context, it is worth noting that the majority of neuroscience-informed models are implemented in simplified simulated environments, not embodied-virtual or robotic agents.

---

2  Theories of emotion generally pertain to one of the following three classes: theories of basic emotions, appraisal theories of emotion, and constructivist emotion theories (including dimentional theories). See Coppin & Sander (2012) for a more detailed description.

## 1.2. Outlook

In our view, despite advances, the literature on emotion modeling suffers from a number of shortcomings. In particular, existing models are often domain specific. For example, models of how emotion can improve communication in social contexts do not address how affective processes can guide environment exploration or behavior regulation when handling conflicting goals, and vice versa. Thus, adapting a model to a new context of application effectively amounts to designing a new model. More importantly, emotion-based architectures are seldom designed as more general-purpose architectures. Overall, the field lacks frameworks for integrating affective processing in intelligent systems that can scale up beyond controlled toy scenarios.

To develop such frameworks, we argue that stronger connections with the neuroscience of emotion is needed. While it is in theory possible to engineer intelligent machines without consideration of living organisms, we believe it is enormously beneficial to take cues from how biology gives rise to intelligent behaviors, as demonstrated by proposals inspired by biological cognition (e.g. Mnih et al, 2015; Cully et al., 2015; Moulin-Frier et al., 2017; Doncieux et al., 2018). As a model of autonomous learning and decision-making, reinforcement learning is a good example of fruitful interaction between neuroscience and artificial intelligence (Neftci & Averbeck, 2019). Interestingly, reinforcement learning has been proposed as a possible framework to bridge the machine learning and emotion modeling communities (Moerland et al., 2018). Indeed, reinforcement learning models acknowledge the central role of affective processing by postulating that reward-seeking drives autonomous behavior. However, this framework primarily addresses processes related to specific forms of learning and decision-making. Notably, it does not encompass processes such as attention and executive control that we believe have strong potential to benefit from affective processing (see Section 3.1). More generally, how affective processes are modeled in robots and artificial agents often contrasts sharply with current knowledge about biological emotion. In the following, we summarize key findings of the neuroscientific literature that highlight the gap between natural and artificial emotion. In particular, we stress the integration between emotion and perception/cognition in humans and animals at multiple levels: brain, body, and behavior.

## 2. Natural emotion: brain, body, and behavior

## 2.1. Emotion and the brain

Historically, the brain basis of emotion was conceptualized in an area-centric manner. For a long period, the hypothalamus was believed to be the emotion center, shifting to the amygdala in the 1980s. In the last decades, not only has the number of regions of the "emotional brain" increased steadily, but how they function via complex circuits is starting to be unraveled. These regions include the medial prefrontal cortex, the orbitofrontal cortex, the cortex of the insula, the thalamus, and many more. Critically, rather than being functionally localized in specific areas, emotion-related processes are implemented by distributed neural circuits that rely on multiple structures at the same time (Pessoa, 2017; Tovote et al. 2015; Lindquist and Barrett, 2012).

More broadly, the classical separation between emotion and cognition has been gradually eroded. Behind the blurring of their boundaries is the notion that mental processes are implemented via large-scale, distributed networks (Sporns, 2010). The networks that have been uncovered in the context of cognitive processes share many nodes (i.e. regions) with those that are important for emotion (Pessoa, 2008; Najafi et al., 2016). Thus, neural computations underlying behavior are implemented via overlapping networks. In this manner, specific brain areas affiliate, or group with, multiple large-scale networks depending on behavioral demands.

Even more generally, the separation between mental domains such as perception, cognition, action, motivation, and emotion, while possibly suitable for a textbook organization, does not reflect the organization of the brain. To understand how the brain generates complex, flexible, and adaptive behaviors it is necessary to understand how brain circuits disrespect standard boundaries. In a very real sense, the domains cannot be separated.

## 2.2. Emotion and the body

Intelligence is not a mere collection of computations occurring in the central nervous system but result from the coupling of the brain, the body, and the environment (Varela et al., 1992; O'Regan & Noë, 2001). From this perspective of embodied cognition, emotion is rooted in homeostatic processes that guarantee bodily integrity, and the associated construction of bodily representations capturing the state of body at any instant. These key functions engage both subcortical and cortical areas. Thus, neuroscientifically grounded theories of emotion attribute a central role to the body in emotional-related processes. For example, in the core affect theory, bodily states are central to emotional experience (Russell, 2003). In the somatic marker theory, associations between particular situations and patterns of elicited physiological and emotional reactions are established, and help shape behavior (Damasio et al., 1996).

## 2.3. Emotion and behavior

Emotion expressions, including those such as facial expressions, gestures, and postures, are an important feature of the relationship between emotion and the body (de Gelder et al., 2015; Cowen et al., 2019). The variety and complexity of processes involved in emotion expression and recognition underlines their importance in human social behaviors.

Emotion-behavior coupling is not limited to communicative functions but is also strongly related to motivation and action generation (Frijda 1986; Blakemore & Vuilleumier, 2017). In living organisms, motivated behaviors are represented in terms of approach and avoidance. Even ostensibly simple behaviors like escape leverage complex cognitive-emotional processes (Evans et al., 2019). More generally, survival – and autonomous function – depends on the ability to generate flexible behaviors and to adapt to dynamical environments. In sum, how an organism acts in its environment is a key problem that depends on emotion-related processes, which therefore is not confined to generating expressive behaviors for communication.

## 3. Toward better models of emotion

Based on the preceding discussion, we propose four principles to motivate guidelines for the development of artificial emotion capable of displaying human-like intelligent behaviors.

## 3.1. Account for emotion-cognition integration

The integration of emotion and cognition in the brain can be used to inform how affective processing interacts with other computational processes, and more generally the notion of architecture modularity. Consider a traditional architecture with standard components such as perception and decision-making (Figure 2A). Recognizing the utility of affective information, models have included an emotion-related component that interfaces with some of its components, for example, allowing motivational factors ("fatigue" and "cold") to influence action selection (Avila-Garcia & Cañamero 2004). Employing affective information in a subset of processes is viable in single-purpose systems that operate in specific, controlled contexts. However, we suggest that more general-purpose architectures should be designed such that emotion-related mechanisms interface with all of the remaining processes. This is because in unconstrained, complex environments systems with *limited computational resources* need effective ways to allocate them and to organize their behavior intelligently. This is where emotion-related processing is important, in particular the concept of value, relevance, and/or significance.

Figure 2B illustrates how all processes that are traditionally separated in distinct modules should actually be more connected and, more importantly, involve emotional processing. But how can this principle be translated into concrete implementation? Consider the example of `attention`, a cognitive operation central to the notion of computational resource allocation. A fruitful way to conceptualize attention is in terms of *priority maps* (Itti et al., 1998). In particular, the priority of a to-be-attended visual item depends on a series of factors, including stimulus salience and top-down control, which can be respectively labeled as perceptual and cognitive factors. Critically, priority also depends on affective and motivational significance, including reward (Anderson and Phelps, 2001; Anderson et al., 2011). For example, an item paired with aversive consequences in the past will acquire negative significance, and gain prioritized processing so that it can be adequately handled. Likewise, an item paired with reward in the past will acquire motivational value. Combined, the determination of the overall object *relevance* integrates multiple factors, typically via competition/cooperation mechanisms (Figure 2C). For example, objects with both high perceptual saliency and high value (e.g paired with reward in the past) can outcompete objects that are goal-relevant. The ability to direct attention according to multiple factors, including affective and motivational ones, is of utmost importance for intelligent systems with finite resources dealing with real-world situations. Note that the proposed approach is different from those using neuromodulation to implement attention-related processes (e.g. Krichmar, 2013), as neuromodulation typically evolves across slower time scales than the "on-the-fly" integration of factors in Figure 2C. The two approaches are complementary.

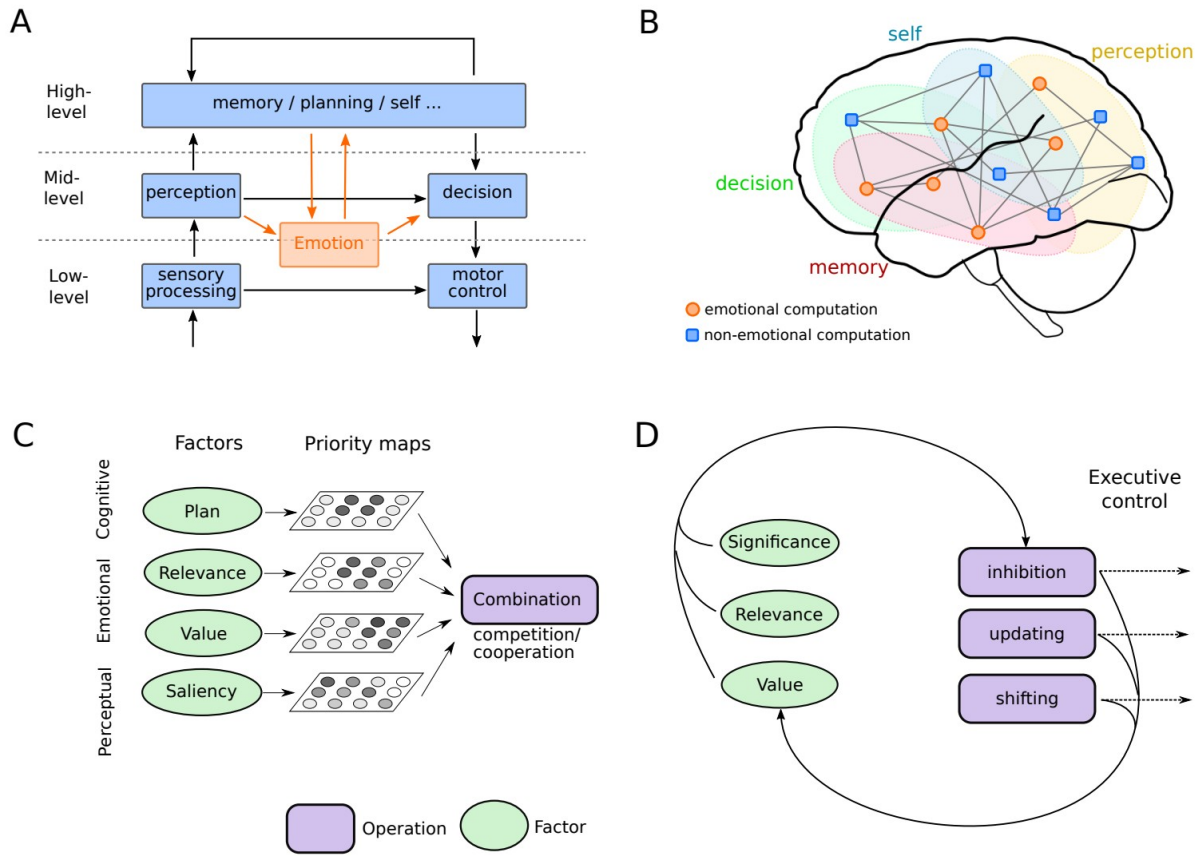

*Figure 2: Emotion integration in robot cognitive architectures. **A)** Traditional cognitive architecture in which emotion interacts with some of the other components. In this type of architecture, emotion is an added module that can be disconnected from others. **B)** Brain-like architecture in which the separation between mental processes is blurred. More importantly, all processes involve a combination of emotional and non-emotional computations. This view emphasizes that affective processing is an integral element of the system. **C)** The process of attention combines multiple priority maps to determine which elements in the environment the robot should attend to. Priority maps are based on multiple factors: perceptual (e.g. based on saliency), cognitive (e.g. based on current plan), but also emotional/motivational (e.g. based on value and relevance). **D)** The process of Executive control is based on a set of operations such as inhibiting, updating and shifting that take into account affective factors to support successful autonomous adaptive behaviors. At the same time, the very same functions help determine factors such as value, relevance, and significance.*

As another example, consider `executive control` (also called "cognitive control"), which includes operations involved in maintaining and updating information, monitoring conflict and/or errors, resisting distracting information, inhibiting prepotent responses, and shifting goals. A useful way to conceptualize executive control is in terms of a set of processes, including inhibition, updating, and shifting (Miyake et al., 2000). Insofar as value, relevance, and/or significance need to be taken into account for proper executive control, emotion/motivation participate in these processes. In other words, objects or contexts that influence cognitive control processes such that rewards (respectively, punishments) ensue, become positively (respectively, negatively) relevant to guide or direct executive control. Why is the architecture in Figure 2A not sufficient? After all, information about what is emotionally/motivationally relevant will be conveyed to the particular architecture components. The central reason is that influences must be *bidirectional* (Figure 2D). For example, dealing with an emotional stimulus or situation requires multiple adjustments, including "updating" to refresh the contents of working memory, "shifting" to switch the current task subgoal, or "inhibiting" to cancel previously planned actions. In this manner, resources are coordinated in the service of proper ("intelligent") function.

## 3.2. Subscribe to principles of embodiment

To stress the importance of embodiment for artificial intelligence, roboticists often use arguments related to morphology and physical interaction with the environment (Brooks, 1991; Pfeifer et al., 2007). Consider a system that must learn the concept of a "chair". Purely vision-based approaches (e.g. using deep neural networks) would need a massive amount of data and would only be able to recognize chairs by shape. In contrast, a humanoid robot able to sit on a flat surface could learn that sitting minimizes energy loss and thus start to learn the functional aspects of chairs. In other words, disembodied machines cannot make sense of the world the same way as humans do. In the context of emotion, we believe that the same reasoning applies. For instance, human-like facial expression recognition should be embedded in a system that can produce expressive behavior and associate it with its own internal states in order to process what is being expressed, by the system itself or others (see also Sagar et al., 2016, on the importance of modeling low-level brain-body mechanisms to produce realistic face animations). Otherwise, it will be little more than a detection device of stereotypical patterns labeled as 'happy' or 'afraid', and thereby unlikely to reach human-level performance.

When addressing emotion embodiment in artificial systems, there has been a focus on how emotion is expressed through the body (e.g. emotion recognition in computer vision, face actuators in social robotics, synthesis of social cues in computer animation). But, for models of emotion and cognition to be more truly embodied, the behavior they implement must be driven by signals related to potential bodily harm, safety, satiation, energy depletion, and so on (Figure 3A; see also Froese & Ziemke, 2009, and Man & Damasio, 2019). In particular, the successful execution of higher-order goals partly depends on the association between a set of actions with their affective and motivational consequences (positive or negative). Therefore, building a robot capable of autonomously and intelligently exploring an unknown environment requires mechanisms to monitor energy level, avoid physical harm, develop a preference for safe locations, attend to objects which are relevant to goals/plans, and switch between goals and behaviors depending on current own and external states (Figure 3B), all of which rely on embodied emotional-cognitive processes.

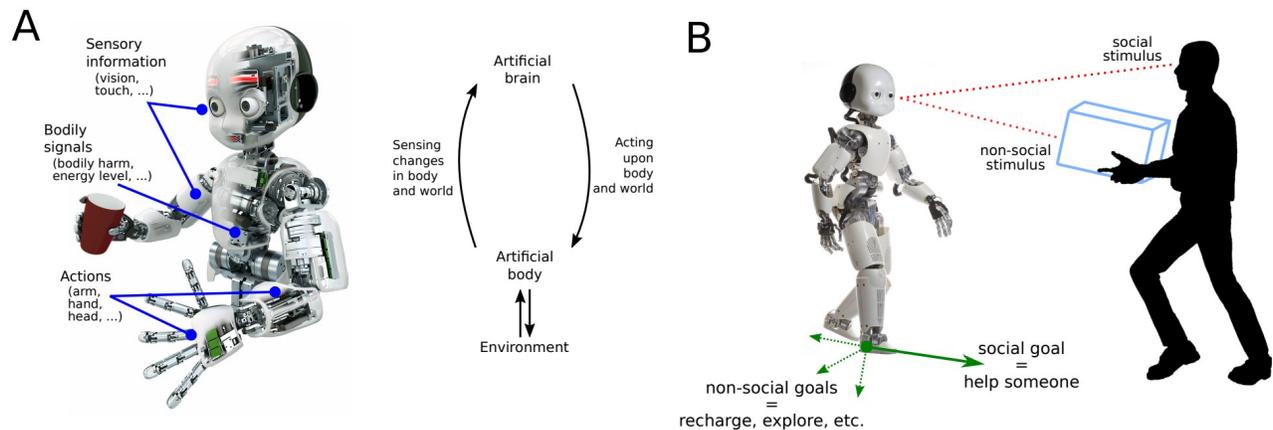

*Figure 3: Embodiment and emotion for intelligent robots. **A)** As embodied intelligent machines, robots are able to acquire information about, and to act upon, the world through a variety of sensors and actuators. The notion of embodiment also includes the processing and regulation of bodily signals related to bodily harm, safety, and energy depletion. **B)** Effective integration of emotion allows robots to generate and coordinate intelligent behaviors in complex situations involving social and non-social interactions. The illustration of iCub robot in A) is reproduced with permission from Antoni Gracia.*

### 3.3. Support both social and non-social behaviors and interactions

Real-world situations consist of a combination of social and non-social interactions with the environment. In order to properly operate in such situations, machines need architectures that can support both types of interactions. In the context of attentional processes, social (e.g. other agents) and non-social (e.g. objects) stimuli can have different types of relevance to the system and should capture attention accordingly. In addition, goals can be social (e.g. helping someone) or non-social (e.g. reaching a location, grasping an object; see Figure 3B). Accordingly, architectures should be able to direct processing and allocate resources in a manner that takes into account both social and non-social priorities. From an engineering perspective, autonomous cars could be safer if they had the capacity to interpret social cues (e.g. pedestrian patterns and interactions); industrial robots could be more efficient if they were able to manage both independent and collaborative tasks; and companion robots could be more engaging and fun if they could develop a "personality" from both social and non-social experiences.

Developing human-like models of emotion for robots has the potential to enable intelligent behaviors that are relevant for both social and non-social contexts. In Section 3.1, we highlighted the importance of integrating emotional factors with attentional processes, which is relevant in non-social situations so that machines can autonomously allocate computational resources to the most important stimuli. But the other side of visual attention is gaze, which is a powerful social cue. Translating a robot's attention into gaze-orienting behaviors provides an intuitive way for signaling to-be-attended objects, preferred objects, or to-be-avoided objects. Belkaid et al. (2017; illustrated in Fig 1) described an approach for top-down modulation of a robot's attention based on emotional signals, such as frustration and boredom, derived from the agent's own evaluation of ongoing performance. Similarly, Broekens & Chetouani (2019) described a model in which emotional expressions rely on reinforcement learning variables (e.g. prediction errors) to define affective states such as joy and distress. Thus, overall, models of emotion can jointly contribute to social and non-social interactions. Returning to the example of collaborative robots, while some aspects of emotion may not be relevant for these applications, this approach would allow the generation of socially relevant behaviors which are grounded in the very same processes that enable them to function autonomously in non-social situations; as opposed to being triggered by ad-hoc modules in a pre-scripted fashion.

## 3.4. Inform research on natural emotion and cognition

Computational and robotic models have the potential to play increasingly important roles in the study of the neural basis of emotion (Arbib & Fellous, 2004; Cañamero, 2019). To do so, machines should be conceived as models which can advance our understanding of human intelligence through the process of recreating it. Can we build machines able to process different types of stimuli and events, to safely explore an unknown environment, to self-regulate and adapt behavior in the face of diverse contexts, to develop long-term knowledge, preferences, goals, and relationships? In doing so, designing intelligent machines can benefit not only *from* but also *to* the study of natural intelligence.

Computational models can inform research on human emotion and cognition at four levels: 1) testing existing theories, 2) proposing new theories, 3) proposing new experiments, and 4) creating opportunities for new experiments (Figure 4). For instance, does the current understanding of how we process social and non-social stimuli (e.g. threatening face vs. snake) suffice to implement similar mechanisms in a robot? Assessment of the current state of knowledge will reveal ambiguities and important gaps in the literature (level 1). For example, how is processing prioritized in the presence of diverse types of distractors (social, non-social, positive, negative)? The process of testing theories should be hypothesis-driven and based on scientific knowledge, rather than solution-oriented (i.e. engineering a functional system) to contribute to the development of new theories (level 2). The process can then suggest new experimental designs to test the validity of the proposed hypotheses (level 3). Finally, modeling intelligent behavior in machines has the potential to lead to innovative experimental research (level 4). For instance, there is a growing body of research investigating aspects of social cognition using robot-based paradigms (Henschel et al., 2020; Belkaid et al., 2021). Indeed, robots offer a unique opportunity to create real-time, yet controlled, interactions.

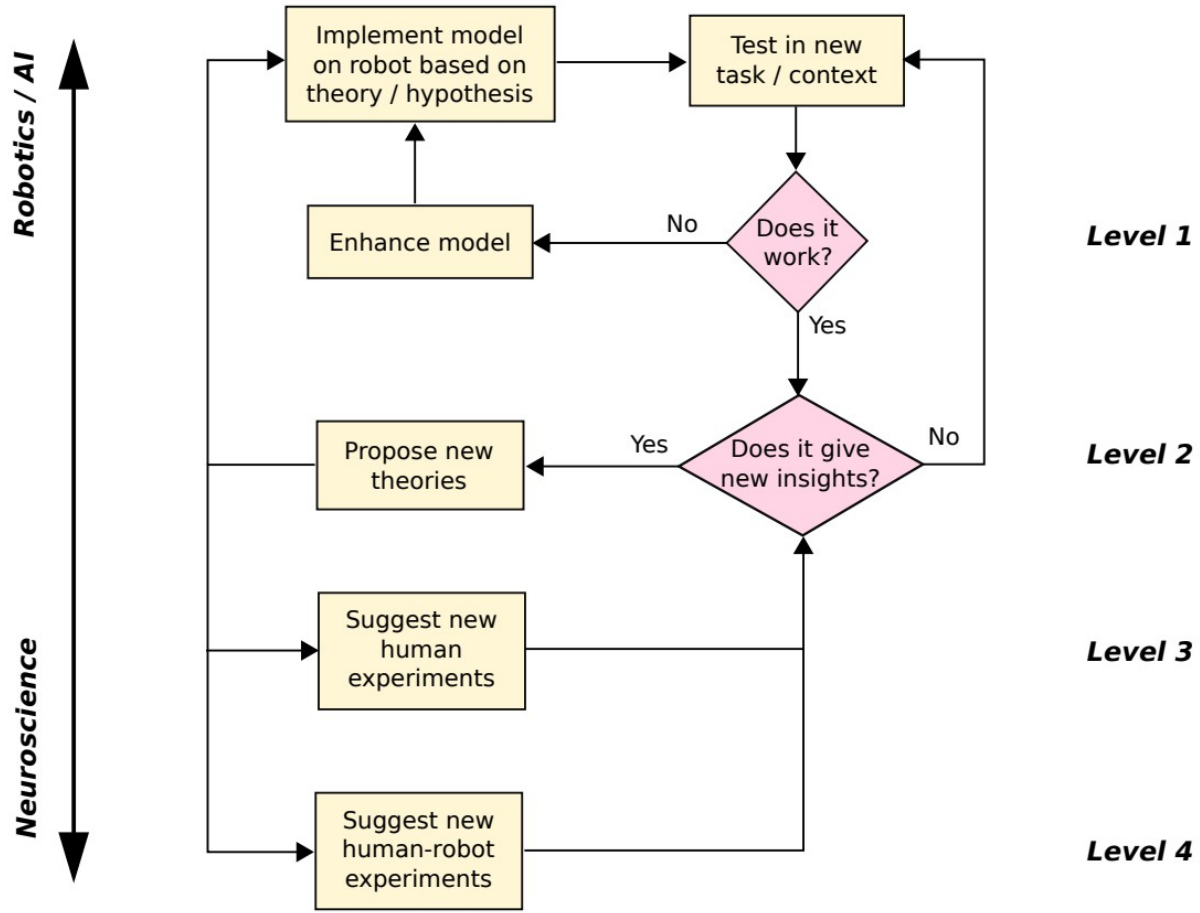

*Figure* 4: Schematic of how emotion modeling in robots can inform neuroscientific research. Levels at which modeling can help advance neuroscientific knowledge: 1) testing existing theories, 2) proposing new theories, 3) proposing new experiments, and 4) creating opportunities for new experiments. Embracing an interdisciplinary approach will be beneficial to both the robotics and the neuroscience communities.

## Discussion

Recent advances in neuroscience emphasize the importance of emotion in human intelligence and stress the interdependent relationship between brain, body, and environment. Modeling and integrating emotion in "cognitive architectures" thus has the potential to help build autonomous social robots able to behave intelligently in diverse and challenging real-world situations. Indeed, several research groups have advocated for interdisciplinary approaches to emotion modeling in machines (e.g. Sloman & Croucher, 1981; Arbib & Fellous, 2004; Cañamero & Gaussier, 2005; Cañamero, 2019). Previous research in social robotics, neurorobotics, and affective computing have made considerable progress in this regard. Yet, modeling human-like emotion in robots remains a challenge.

In this paper, we provided an overview of the state of the art in emotion modeling. We attempted to summarize the research landscape from various disciplines in terms of five criteria: embodiment, behavior, architecture design, theoretical approach, and research goal. By linking this literature to current knowledge about human emotion, we identified a set of potential issues that can be summarized as follows. On the one hand, emotion-based robotic architectures are often domain-specific, where emotion-related mechanisms are restricted to specific processes or modules. As a result, these models fail to capture fundamental aspects of affective processes and, importantly, to account for their multiple roles across diverse functions. On the other hand, neuroscience-informed models of emotion that describe affective processes more accurately are generally tested via simulations using rather simplified inputs; they have not been implemented on embodied machines that interact with the physical and social worlds. We hope our proposal contributes to the development of research guidelines for designing autonomous social machines in a manner that is centered on the integration between emotion and cognition.

A questions this paper might raise is 'how much human-like emotion do we want machines to have?'. Our goal is to encourage a focus on "affective processing" that is separate from the subjective experience of emotion per se. We propose that key features of affective processes are constitutive and inseparable from human-like intelligence, and that modeling those features is essential if we want to build autonomous social robots able to tackle real-world problems. Moreover, we argue that the process of seeking to model human-like emotion in machines following an interdisciplinary approach has the potential to unravel fundamental computational principles that may be difficult to capture through current neuroscientific methods. Further understanding those principles would help answer the question of whether their implementation in machines is ultimately desirable for human society.

## Declarations

**Data and materials availability**  All data needed to evaluate the conclusions in the paper are present in the paper.

**Funding**  No funding was received to assist with the preparation of this manuscript.

**Competing interests**  The authors have no relevant financial or non-financial interests to disclose. MB is co-Guest editor of the Special issue. The authors declare no conflict of interest.